\documentclass[twoside]{article}
\usepackage[accepted]{aistats2018}
\usepackage{amsmath}
\usepackage{amsfonts}
\usepackage{url}
\usepackage{natbib}

\usepackage{graphicx}
\usepackage{subcaption}
\usepackage{booktabs}

\renewcommand{\vec}[1]{{\boldsymbol{\mathbf{#1}}}} 

\newcommand{\R}{\mathbb{R}}

\newcommand{\set}[1]{\mathcal{#1}}

\newcommand{\grad}{\nabla}
\newcommand{\sample}{\sim}
\newcommand{\given}{|}
\newcommand{\CDF}{\textrm{CDF}}

\newcommand{\normal}{\mathcal{N}}
\newcommand{\norminv}{\Phi^{-1}}
\newcommand{\chisq}{\chi^2}

\newcommand{\E}{\mathbb{E}}
\newcommand{\var}{\textrm{Var}}
\newcommand{\MI}[2]{\textrm{MI}(#1;#2)}
\newcommand{\median}{\operatornamewithlimits{median}}

\newcommand{\GR}{\textrm{GR}}
\newcommand{\G}{G}
\newcommand{\EFF}{\textrm{EFF}}
\newcommand{\ESS}{\textrm{ESS}}
\newcommand{\RESS}{\textrm{RESS}}
\newcommand{\NESS}{\textrm{NESS}}
\newcommand{\ESSD}{\textrm{ESSD}}

\newcommand{\target}{{p^\star}}
\newcommand{\utarget}{{\tilde{p}}}
\newcommand{\q}{q}

\newcommand{\dimx}{d}
\newcommand{\Dimx}{D}
\newcommand{\nsamples}{N}
\newcommand{\nchains}{k}
\newcommand{\Nchains}{K}

\newcommand{\KS}{\textrm{KS}}

\newcommand{\sampler}{S}
\newcommand{\targetset}{\set{M}}

\newcommand{\approach}{\stackrel{d}{\rightarrow}}

\begin{document}

\twocolumn[

\aistatstitle{How well does your sampler really work?}

\aistatsauthor{Ryan Turner \And Brady Neal}

\aistatsaddress{MILA, Universit\'{e} de Montr\'{e}al \And MILA, Universit\'{e} de Montr\'{e}al}]

\begin{abstract}
We present a new data-driven benchmark system to evaluate the performance of new MCMC samplers.
Taking inspiration from the COCO benchmark in optimization, we view this task as having critical importance to machine learning and statistics given the rate at which new samplers are proposed.
The common hand-crafted examples to test new samplers are unsatisfactory; we take a meta-learning-like approach to generate benchmark examples from a large corpus of data sets and models.
Surrogates of posteriors found in real problems are created using highly flexible density models including modern neural network based approaches.
We provide new insights into the real effective sample size of various samplers per unit time and the estimation efficiency of the samplers per sample.
Additionally, we provide a meta-analysis to assess the predictive utility of various MCMC diagnostics and perform a nonparametric regression to combine them.
\end{abstract}

\vspace{-5mm}
\section{INTRODUCTION}
\vspace{-1mm}

Markov chain Monte Carlo (MCMC) methods have seen a huge increase in use over the last few decades.
The goal in MCMC methods is to take samples from a complex probability distribution $\target$ given access only to its unnormalized density $\utarget$.
The primary use case for MCMC methods is sampling from Bayesian posteriors for the purpose of Monte Carlo integration.
These posteriors are intractable to normalize and sample from in complex models.

Approaches such as rejection sampling provide exact independent samples, and importance sampling provides exact independent (but weighted) samples.
These approaches are generally computationally inefficient (rejection sampling) or are statistically unsound (importance sampling) except in very low dimensional problems~\citep[Ch.~29]{Mackay2003}.
MCMC methods produce a Markov chain that marginally samples from the target distribution $\target$ exactly and have a low per sample computation cost.
The downside is that they provide a sequence of \emph{correlated} samples, albeit marginally from the target distribution.
Therefore, any estimates derived from an MCMC chain of length $\nsamples$ will have far less accuracy than $\nsamples$ iid samples.
Despite there being numerous MCMC diagnostics, there is no practical way to guarantee the accuracy of derived estimates in practice.

Each machine learning conference contains a publication proposing a new variation on MCMC methods.
The community lacks a method to determine if these new methods actually sample from posteriors found in real problems with improved accuracy over existing samplers.
New methods are benchmarked either 1)~via hand-crafted toy problems (where a ground-truth is known) or 2)~via test set performance on real problems.
The issue with hand-crafted examples is obvious: Performance on these problems may have little relation to performance on real problems; and it is at odds with accepted practice in modern machine learning.
Benchmarking via test set performance on real problems is laudable.
However, it confounds the specification of the model and priors with the performance of the sampler.
In a misspecified model it is possible that a sampler stuck in an unrepresentative part of the posterior could actually have higher test set performance.
Conversely, a better sampler may improve test set performance by having good local mixing; however, it is still nowhere near exact iid samples.
There is no way to quantify the distance to exact iid samples from test set performance alone.

Whether current samplers are providing samples from anything close to the true posterior on difficult problems is of critical importance for determining future research directions.
Are samplers with higher test set performance actually sampling from real posteriors more faithfully?
Can we sample with any fidelity from complex high dimensional distributions?
Is that merely a ``fool's errand''?
The answers to these questions will determine if it is a worthwhile endeavor to continue to hone MCMC methods for application in successful modern models such as deep neural nets.

Practitioners in Bayesian statistics have long faced the dilemma of whether they can trust the output of their sampler, in particular, because statisticians are not traditionally concerned only with test set error rates.
As a result, there have been decades of work in developing MCMC diagnostics that aim to \emph{alert} a practitioner to a poorly behaving chain~\citep{Cowles1996}.
This is often described by whether their chain is \emph{mixing well}.
In other words, if a chain has a long auto-correlation time, the entire chain may be of equivalent accuracy to just a few iid samples.
The diagnostics, by construction, have a low type I error: That is, if a chain closely resembles iid samples, they will not alert that it is mixing poorly.
However, there are no guarantees on type II error: If a chain is mixing poorly, will the diagnostic alert?
Indeed, there are many ways to construct examples where an MCMC procedure undetectably fails:
distant modes, Neal's funnel~\citep{Thompson2011}, extreme ill-conditioning, etc.
However, are these realistic stress tests for MCMC methods or merely pathological cases?
We do not know.

We propose a new data-driven approach to create a benchmark that estimates how well various MCMC procedures work on real problems.
Arguably, algorithms in machine learning and statistics rely on the ``workhorses'' of either optimization or sampling methods.
The world of (non-convex) optimization has already tackled this challenge with the COCO benchmark~\citep{Hansen2016}, which contains a test battery of difficult optimization problems.
Various approaches are tested to validate if they can optimize the objective function to a target level within a fixed number of function evaluations.
Our approach is an analogous system for sampling methods.
However, we aim to further improve upon this using flexible (including neural net based) benchmark examples that have been trained to match posteriors found in practice.

In our approach we use a large ``data set of data sets'' and a diverse ``model zoo'' to create a representative set of examples.
Long MCMC chains are drawn (using NUTS~\citep{Hoffman2014}) from each of these posteriors.
Flexible unsupervised models that serve as a \emph{ground-truth} in the benchmarking phase are fit to the chains to construct the benchmark examples.

More concretely, each combination of real data set (e.g., MNIST) and real model (e.g., logistic regression) results in a Markov chain from NUTS\@.
We then fit an unsupervised model (e.g., mixture of Gaussians) to this chain to serve as a \emph{benchmark example distribution}.
Once trained, these benchmark example distributions are functionally equivalent to hand-crafted examples such as the toy posterior distributions usually used to benchmark samplers (or such as those in COCO)\@.
However, these examples are not hand-crafted but, rather, are much more representative of real problems.
Because it is possible to draw exact (iid) samples from the benchmark example distributions, we now have a ground-truth set of samples to validate the accuracy of the sampling methods.
We derive a variety of metrics to summarize the performance of a sampler to compare its output to ground-truth iid samples.
The ground-truth samples also allow us to assess how well the MCMC diagnostics actually predict estimation performance.
In particular, we look at the effective sample size (ESS) because it provides a concrete statement on the quality of an MCMC chain~\citep{Kass1998}.

\paragraph{Contributions}
We summarize the contributions of this work as follows:
1)~We provide a new and novel benchmark to describe how well various samplers work on realistic problems.
This involves design of fair and sensible metrics to score samplers across problems.
This system will be provided as a software system that will serve as a practical tool in algorithm development analogous to MLcomp/CodaLab or COCO\@.
2)~We shed light on how well the common MCMC diagnostics predict the real estimation performance of MCMC methods.
We further create a data-driven meta-diagnostic by combining various MCMC diagnostics to predict the real performance of a sampler.

\paragraph{Related work}
The closest existing system is SamplerCompare of~\citet{Thompson2011}, which tests samplers on a handful of hand-crafted stress-test cases such as Neal's funnel.
However, SamplerCompare is more an R package to aid sampler evaluation than a complete benchmark.
A recent piece of work from systems biology~\citep{Ballnus2017} compares various samplers for dynamical systems (i.e., filtering) on a set of hand-crafted ODE systems inspired by biological applications.
Sampling in ODE systems is not very representative of the challenges in sampling from posterior distributions in machine learning models.

\section{BACKGROUND}

\begin{figure*}
    \centering
        \includegraphics[width=0.8\textwidth]{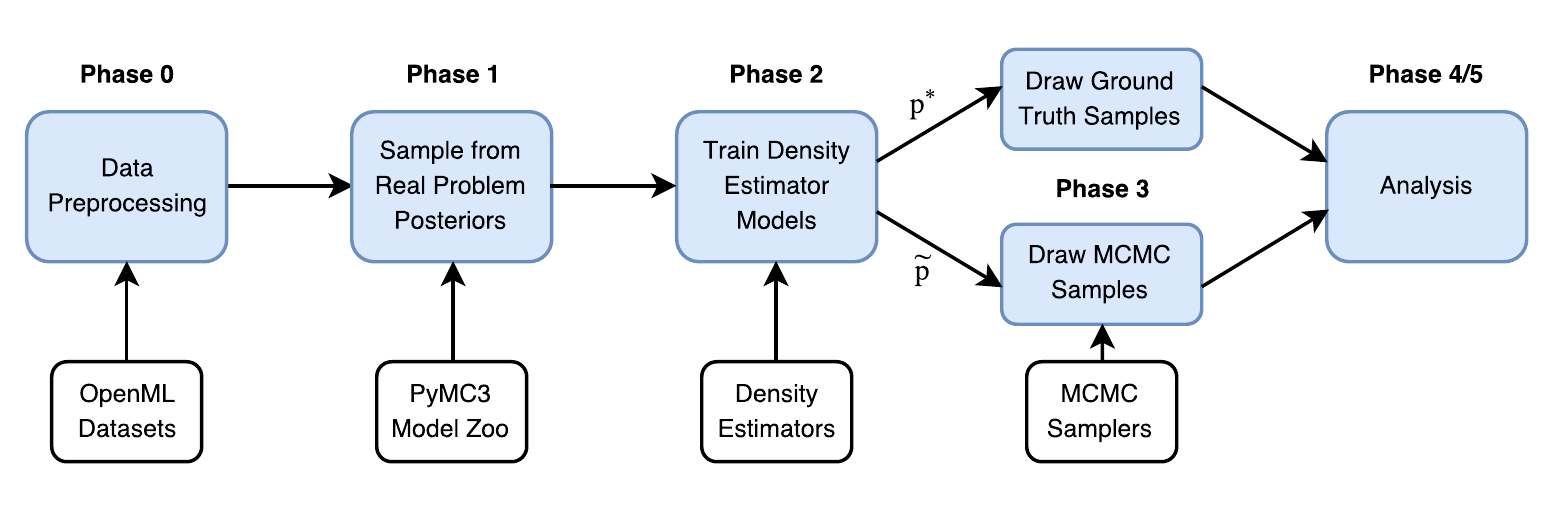}
    \caption{{\small Flowchart illustrating the six phases in our methodology.
    Phases 0--2 are for creating benchmark examples and are not re-run when new samplers are tested.
    Phase 2 includes mixture models and modern neural net methods.}}
    \label{fig:flowchart}
\end{figure*}

The notion of a black box is highly relevant to conceptually understanding this work.
Fundamentally, an MCMC sampler is a system that takes a black box that computes an unnormalized density $\utarget \propto \target$ (and possibly its gradient $\grad \log\utarget$) and a previous sample $\vec x_{t-1} \in \R^\Dimx$ in the Markov chain; it outputs another sample $\vec x_t \in \R^\Dimx$.
Once the Markov chain has converged, these samples are theoretically guaranteed to marginally come from the density $\target$, albeit with temporal correlation.
If the previous sample was drawn exactly, $\vec x_{t-1} \sample \target$, then $\vec x_{t} \sample \target$ exactly as well.
This condition is a result of \emph{detailed balance}.

By analogy, optimization algorithms take an objective function $f \in \R^\Dimx \rightarrow \R$ (and possibly its gradient $\grad f$) as a black box and produce points $\vec x_t \in \R^\Dimx$ that successively minimize $f$ as much as possible.
Just as COCO provides its benchmark objective functions $f$ as a black box to the optimizers and keeps hidden the true optimum, our benchmark provides the unnormalized density $\utarget$ as a black box to the samplers.
Our benchmark keeps hidden the parameterization of $\utarget$ needed to efficiently take iid samples from $\target$.

\subsection{Traditional MCMC Diagnostics}

Given that we have a ground-truth to evaluate the performance of the various samplers, we can also benchmark the diagnostics by seeing how predictive they are of actual performance.
In particular, we consider three diagnostics in this paper: ESS, Gelman-Rubin (GR), and Geweke.
ESS aims to estimate how many iid samples have the same estimation performance as the correlated samples found in the MCMC chain.
Gelman-Rubin~\citep{Gelman1992} and Geweke~\citep{Geweke1992} more closely follow a test statistic paradigm than an estimation one.
Gelman-Rubin compares the variance within a single chain to variance between chains (independent restarts)\@.
This quantity should be close to one for well-mixing chains and can be very large for poorly performing chains.
The Geweke diagnostic works on a single chain and compares the variance between different chunks.

The ESS diagnostic is basically a rescaling of the expected square error (i.e., MSE) on estimating the mean in a \emph{single dimension} (marginal) of $\vec x$.
ESS is based on the notion that for the marginal $x_\dimx$:
\begin{align}
  \E_\target[({\hat{\mu}}_\dimx - \mu_\dimx)^2]
  &= \var_\target[{\hat{\mu}}_\dimx - \mu_\dimx] + \E_\target[{\hat{\mu}}_\dimx - \mu_\dimx]^2 \nonumber \\
  &= \var_\target[x_\dimx]/\nsamples\,, \quad \dimx \in 1\!\!:\!\!\Dimx\,, \label{eq:sq loss} \\
  {\hat{\vec\mu}} &:= {\scriptstyle{\frac{1}{\nsamples}\sum_{i=1}^\nsamples \vec x_i}}\,, \quad \vec\mu := \E_\target[\vec x]\,,
\end{align}
which utilizes that ${\hat{\mu}}_\dimx$ is an unbiased estimate of $\mu_\dimx$.
We are careful to distinguish expectations and variances with respect to $\target$, where $\vec x$ is iid, from $\q$ where the samples are correlated and from an MCMC method.
Naturally, by re-arranging~\eqref{eq:sq loss}, the effective sample size for non-iid samples is:
\begin{align}
  \ESS := \frac{\var_\q[x_\dimx]}{\E_\q[({\hat{\mu}}_\dimx - \mu_\dimx)^2]} \in \R^+\,,
  \label{eq:def ESS}
\end{align}
which unlike~\eqref{eq:sq loss} can be estimated without ground-truth samples from $\target$.
However, the difficult denominator term is typically estimated using the empirical linear auto-correlation of the Markov chain.
This linearity assumption is obviously a potential source of error in the $\ESS$.
The fixation in estimating the accuracy of the mean $\hat{\mu}$ is also a weakness.
In Section~\ref{sec:analysis}, we look at the \emph{real} effective sample size by comparing estimates with the ground-truth samples.
It also allows us to look at other measures than simply the fidelity in matching the means ($\hat{\mu} - \mu$).

\section{METHODOLOGY}

Our benchmark system follows a six phase approach, which we explain at a high level in this section.
In Section~\ref{sec:details}, we provide low-level specifics.
A graphical summary of this section is provided in Figure~\ref{fig:flowchart}.

In phase~0, we create a ``corpus'' of data sets that we refer to as a ``data set of data sets.''
This is meant to create a realistic sample of problems that a practitioner may encounter ``in the wild.''
Such an approach was also taken in the AutoML competition~\citep{Guyon2015} and the automated statistician project~\citep{Lloyd2014}.
Our approach can be thought of as a form of \emph{meta-learning}~\citep{Vilalta2002}.

In phase~1, we use a model zoo to simulate a variety of (Bayesian) models that a practitioner might attempt to apply to a real problem.
There are models for regression and classification.
Each model/data set pair results in a posterior over a parameter space, which varies in dimensionality depending on the problem.
Except in very simple cases (e.g., linear regression), we are not able to obtain samples from these posteriors exactly.
We use NUTS, the default sampler in probabilistic programming languages (PyMC3~\citep{Salvatier2016} and Stan~\citep{Carpenter2016}), because it is generally considered to be a good off-the-shelf sampler especially when paired with the intelligent initialization and automatic tuning found in these systems.
Therefore, by running multiple long chains of NUTS on the posteriors, we obtain a sufficient approximation and representation for passing to phase~2.

In phase~2, we run various density estimation models to generate benchmark example distributions on the Markov chains from phase~1.
We run a separate training procedure on each model/data set pair.
These benchmark example distributions serve as surrogates for the real posteriors found in phase~1.
Note that the goal is not to replicate the posteriors from phase~1 exactly, but to generate example distributions that are \emph{qualitatively similar} to the real posteriors in phase~1.
This gives us example distributions that are more realistic than the usual hand-crafted toy problems.
Nonetheless, we train multiple models and take the one with the highest held-out likelihood on the last 20\% of the Markov chain found in phase~1.
We use held-out likelihood as it is the most widely accepted generic method of verifying model fidelity.
Model checking diagnostics are also run to verify the similarity between the benchmark example distributions (surrogates) and their corresponding Markov chains from the real posteriors (originals)\@.

When selecting models for use as benchmark example distributions in phase~2, we have the following requirements:
1)~The models are flexible enough to closely fit the posteriors found in phase~1.
2)~They can serve as a black box, providing an unnormalized density $\utarget$ (and its gradient) when queried at an arbitrary point~$\vec x$.
3)~We can efficiently sample (ground-truth) from them given their parameters (which are hidden from the samplers)\@.

In phase~3, we benchmark a collection of samplers.
If someone invents and provides a new sampling algorithm, it is added in phase~3.
Phases 0--2 remain fixed as new samplers are submitted to be benchmarked.
Each sampler to be benchmarked is run on each of the benchmark example distributions for multiple chains.
Each chain is allowed to run for a fixed period of time.
The raw samples from these Markov chains are saved as the output of phase~3.

In phase~4, we take a large number (e.g., $\sim \!10^5$) of exact iid samples from the benchmark example distributions as a ground-truth.
The square loss between point estimates (e.g., ${\hat{\mu}}_\dimx$ or ${\hat{\sigma}}_\dimx^2$) taken from the Markov chains from phase~3 and the point estimates from the exact chains are aggregated.
We also compute and store the MCMC diagnostics for each chain.

In phase~5, we aggregate the performance results by looking at the real effective sample size as derived from the square errors in point estimation.
We also define transformations of the real effective sample size, which we will refer to as efficiency, normalized effective sample size, and effective sample size deviation.
In addition, we perform a meta-analysis using Gaussian process (GP)~\citep{Rasmussen2006} regression to predict the real effective sample size given the MCMC diagnostics.
This will be useful to practitioners aiming to quantify their confidence in an MCMC-based estimate using the diagnostics available.

\section{ADDITIONAL DETAILS}
\label{sec:details}

In this section we present additional details for the construction of each phase.

\subsection{Phase~0: A Data Set of Data Sets}

Phase~0 involved downloading 2,200 data sets from \url{openml.org} to form our data set of data sets.
We considered other sources, such as the classic UCI repository, \url{mldata.org}, and Kaggle, but settled on OpenML because it had the most standardized format and consistent meta-data.
Such systems are necessary for automated processing.

The data sets were diverse in that their dimension varied from 1 to 61,359, sample size from 5 to 7,619,400, and the number of output classes in classification problems varied from binary to 100.

After downloading, we subjected each data set to some preprocessing to simulate the diverse set of practices a practitioner might follow.
Each data set was randomly preprocessed in one of three ways: standardization, robust standardization (using medians and inter-quartile ranges), or whitening.
Categorical variables were represented with one-hot encodings.

\subsection{Phase~1: Sampling from the Model Zoo}

For the model zoo, we used all of the standard models (regression and classification) typically used with PyMC3.
This includes generalized linear models (GLMs) such as logistic regression, but also less ``vanilla'' GLMs such as robust linear regression (linear regression with Student's-$t$ noise)\@.
In addition to models that are linear in the feature space, we included models that are linear in a second order transformation of the feature space.
We included Gaussian processes with unknown hyper-parameters (e.g., MCMC sampling was done on the unknown hyper-parameters)\@.
Bayesian neural networks were also included.

To keep compute time reasonable, we limited the sample size for expensive models (e.g., GPs), and placed some limits on input dimensionality.
Where dimensionality needed to be reduced we used PCA~\citep{Jolliffe1986} as that is the most frequently used method in practice to reduce dimensionality.

\subsection{Phase~2: Fitting Flexible Surrogates}

There are three varieties of models that satisfy the requirements (flexibility, tractable density, and fast exact sampling) for benchmark example densities: mixture models, RNADE~\citep{Uria2013}, and Real NVP~\citep{Dinh2016}.
In each example, we pick the model that has the highest held-out likelihood on the last 20\% of the chain.

For mixture models, we considered mixture of Gaussians (MoG) with expectation-maximization (EM)~\citep{Dempster1977} and variational MoG\@.
Note that, for simplicity, these models are \emph{not} themselves fit using MCMC\@.
The Bayesian Occam's razor effect~\citep{Jefferys1992} allowed us to simply fix the number of mixture components to 25 in the variational MoG\@.
We used five-fold cross-validation to select the number of components in the EM MoG\@.
There is no consistent winner between these models, so which one we use depends on the example.

We also tuned the RNADE learning rate and hyper-parameters based on pilot runs.
Surprisingly, the mixture models often, but not always, out performed RNADE on the held-out likelihood.
Real NVP based models struggled to achieve competitive test set scores.

These models behave better numerically when trained on standardized data.
Care is taken to reverse this standardization in phase~3, so the samplers are forced to attempt to sample from the posterior in its original scale, which is more challenging.

\subsection{Phase~3: Running the Samplers}

Phase~3 forms the real ``meat'' of the benchmark.
This is where candidate sampling algorithms are actually run on the benchmark example densities.

Whether originally designed this way or not, nearly all respected MCMC procedures proceed by proposing a new point using a \emph{proposal distribution} which is then accepted or rejected using a Metropolis-Hastings step.
Therefore, the difference between samplers is based upon their proposal distributions.
We provide a preview of the proposals used in Section~\ref{sec:results}.

The most widely used, until recently, MCMC procedure was \emph{random walk Metropolis}, which uses a Gaussian random walk proposal
$p(\vec x_t \given \vec x_{t-1}) = \normal(\vec x_t \given \vec x_{t-1}, \Sigma)$, where $\Sigma$ is typically diagonal.
Modern packages such as PyMC3 allow for automatic tuning of the proposal width $\Sigma$, which is critical to achieve good performance.
We also consider Cauchy and Laplace distributed proposals for comparison.

We include Hamiltonian Monte Carlo (HMC)~\citep{Duane1987} methods, which also utilize gradient information to more efficiently ``explore'' the space.
Recently, the No-U-Turn-Sampler (NUTS)~\citep{Hoffman2014} was introduced as an extension of HMC that automatically adapts some of its tuning parameters in order to attempt high off-the-shelf performance.
We include an alternate auxiliary variable method known as \emph{slice sampling}~\citep{Neal2003}, which we apply in a coordinate Gibbs-like fashion.

We also alternate different proposals to form compound proposals.
For instance, we consider mixing expensive efficient proposals like NUTS with cheap inefficient proposals like random walk Metropolis.

Finally, we consider an unconventional sampler known as \emph{emcee}~\citep{Foreman2013} which is popular in fields such as astrophysics, but 
has not gained much use in machine learning.
It works by running multiple ``walkers'' to explore the space in parallel.
Emcee is very fast and can be parallelized, but its efficacy in higher dimensions is somewhat controversial.

\paragraph{Initialization}
The accuracy of MCMC based estimates are a function of two factors: the \emph{burn-in time} and \emph{the mixing time}.
Burn-in time, or time until convergence, is how many steps $k$ are required before $p(\vec x_k) \approx \target$ if $\vec x_0 \sample p_0$, where $p_0$ is some distribution to initialize the chain.
The mixing time, or memory length, is how long it takes to get an independent sample once a chain has converged: how many steps $k$ are required before $\MI{\vec x_k}{\vec x_0} \approx 0$ if $\vec x_0 \sample \target$.
The burn-in time is crucially dependent on the initialization while the mixing time is purely a function of the proposal.

In order to evaluate these two effects separately, we offer two options for initialization:
1)~initialize the chain from an exact sample (because we can do that with the benchmark density examples), or
2)~initialize from an ADVI~\citep{Kucukelbir2017} fit to the example density.
Additionally, most methods benefit from a prior guess at the relative scale of the variables before tuning.
We can use the resulting scales from ADVI for this purpose as well.
We use the PyMC3 defaults for these tuning parameters as that is what a practitioner is most likely to use in practice.
However, alternate schemes can certainly be used within the benchmark.

\subsection{Phase~4: Performance Aggregation}

Each sampler is run for a fixed time limit of 15 minutes of CPU time.
We log the performance of the chain along a uniform grid of 100 points in time (i.e., every 9s) to monitor \emph{real} convergence over time.
Fair evaluation requires evaluating each sampler with a fixed time budget rather than a fixed number of samples.
We expect samplers such as NUTS to be very efficient and high performing on a per-sample basis.
However, they require significantly more computation (including gradients) per sample than simpler methods.
Therefore, their comparison is not as obvious a-priori.
We also log the traditional MCMC diagnostics of each chain.

\subsection{Phase~5: Analysis}
\label{sec:analysis}

To summarize the performance of a Markov chain in comparison with ground-truth samples we need to define some evaluation quantities.
First, recall that we have $\Nchains$ Markov chains
$\{\vec x_{1:\nsamples_\nchains}\}_{\nchains=1}^\Nchains$
for each example $\target \in \targetset$ and sampler $\sampler \in \set{\sampler}$.

Each sampler is evaluated on each example separately and can be scored relative to a variety of \emph{estimators} $\hat{\theta}(\vec x_{1:\nsamples})$.
Analogous to~\eqref{eq:def ESS}, we can score the samples of a Markov chain by the closeness of its mean on a dimension $\dimx$ to the ground-truth samples: $\theta = \E[x_\dimx]$ and
\smash{$\hat{\theta}(\vec x_{1:\nsamples}) = \frac{1}{\nsamples}\sum_{i=1}^\nsamples [\vec x_i]_\dimx$}.
We can also consider how close the variance of the Markov chain samples match the ground-truth samples: $\theta = \var[x_\dimx]$.
This flexibility is a generalization of ESS\@.
As in~\eqref{eq:def ESS}, we assume the estimators \smash{$\hat{\theta}$} are unbiased, and just as with the sample mean $\hat{\mu}$: $\var_\target[\hat\theta] \propto \nsamples^{-1}$.
Furthermore, we assume here that each dimension of the samples $\vec x$ has been standardized using the variance of the ground-truth samples, which makes the estimation errors on each dimension $\dimx$ comparable even when their units differ.

\begin{figure*}
    \centering
    \begin{subfigure}[b]{0.3\textwidth}
        \includegraphics[width=\textwidth]{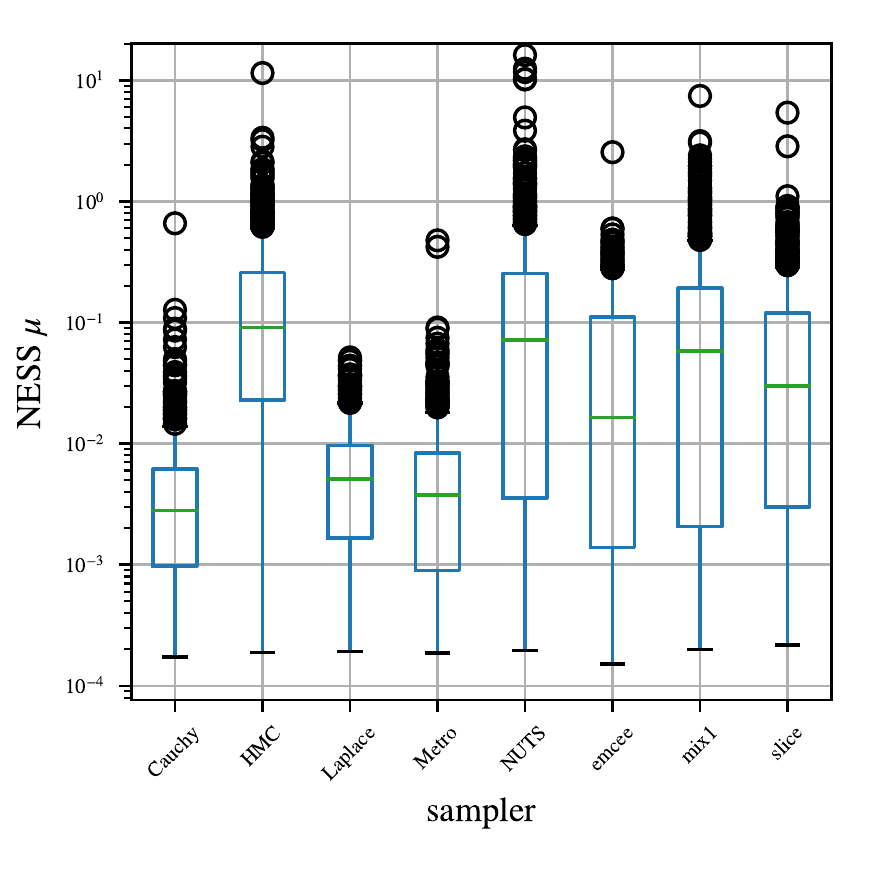}
    \end{subfigure}
    \begin{subfigure}[b]{0.3\textwidth}
        \includegraphics[width=\textwidth]{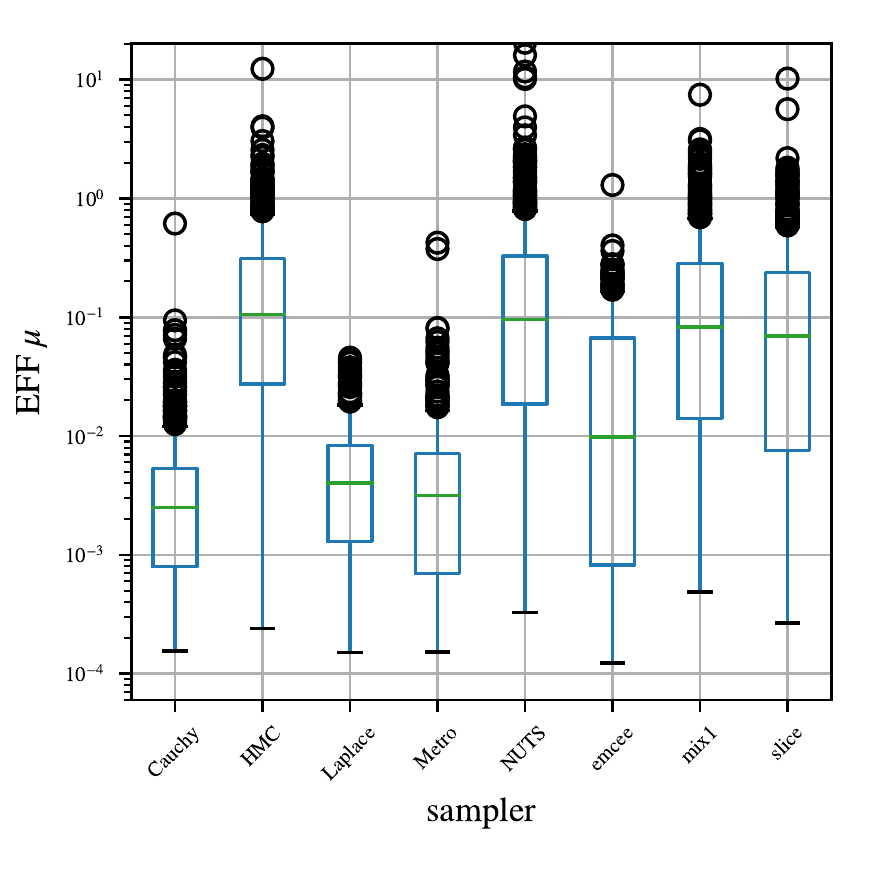}
    \end{subfigure}
    \begin{subfigure}[b]{0.3\textwidth}
        \includegraphics[width=\textwidth]{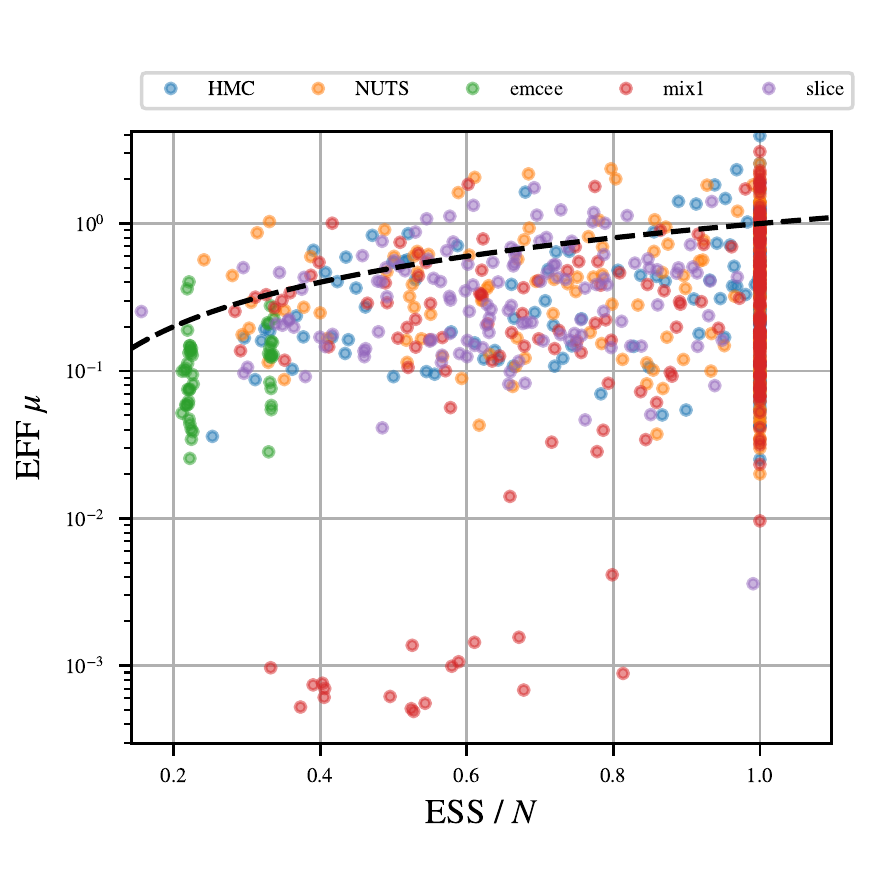}
    \end{subfigure}
    \caption{{\small
  Performance summaries:
    The box plots demonstrate the distribution on NESS (left) and efficiency (center) conditional on the sampler achieving an RESS of at least 12 to only show the mode where the samplers don't completely fail.
    We also show a calibration plot to assess if ESS is a good predictor of efficiency with the diagonal in dashed black.
    Cauchy and Laplace refer to random walk Metropolis with these corresponding proposals.}}
    \label{fig:box}
\end{figure*}

\paragraph{Real ESS}
In analogy to the ESS diagnostic we define the \emph{real ESS} (RESS) based on the estimation error relative to the ground-truth:\footnote{For brevity, we
simply write $\RESS$ rather than $\RESS_{\sampler,\target}$ to denote the RESS of sampler $\sampler$ on example $\target$.
The same applies to the other performance metrics.}
\begin{align}
  \RESS &:= \frac{R}{\textrm{mean sq.~error}}
  = \frac{R \Nchains}{\sum_{k=1}^\Nchains ({\hat{\theta}}_k - \theta)^2} \in \R^+\,, \nonumber \\
  R &:= \E_\target[({\hat{\theta}} - \theta)^2] = \nsamples \var_\target[\hat{\theta}]\in \R^+\,,
  \label{eq:def RESS}
\end{align}
where $\Nchains$ is the number of independent MCMC chains and $R$ is a constant to make RESS comparable across different types of estimators $\hat{\theta}$; it also ensures that $\RESS$ tends towards $\nsamples$ when the samples are iid.
We do not need the $\var[x]$ term from~\eqref{eq:def ESS} because the samples have been standardized using the ground-truth samples' scale.
If the estimator $\theta$ in~\eqref{eq:def RESS} is the mean $\mu_d$ then $R=1$.
Likewise, $R=2$ for variance $\sigma^2_d$ estimation, provided $N$ is large enough for the sample distribution on $\sigma^2_d$ to be approximately Gaussian.

We also consider the Kolmogorov-Smirnov (KS) distance between the samples and the ground-truth samples as a metric.\footnote{Recall
that the KS distance between samples $x_{1:\nsamples}$ and a CDF $F$ is given by $\max_a |\hat{F}(a) - F(a)|$ where $\hat{F}$ is the empirical CDF on $x$.}
This also results in a separate metric on each marginal.
To match the $N^{-1}$ convergence assumption of~\eqref{eq:def RESS} we use
$\sum_{k=1}^\Nchains \KS_\dimx(\vec x_{1:\nsamples}^k, \target)^2$ as the denominator in~\eqref{eq:def RESS}, where $\KS_\dimx$ signifies the $\KS$ distance on the marginal $x_\dimx$.
By numerically integrating~\eqref{eq:def RESS} with the Kolmogorov distribution, one finds that $R = 0.822$ for the KS metric.

RESS is also general in that we can sensibly combine the errors across dimensions by evaluating multivariate estimators $\hat{\vec\theta} \in \R^D$:
\begin{align}
  \RESS = \frac{R \Nchains D}{\sum_{k=1}^\Nchains ||{\hat{\vec \theta}}_k - \vec\theta||_2^2} \in \R^+\,,
  \label{eq:def MV RESS}
\end{align}
assuming that $\hat{\vec\theta}$ is an unbiased estimator of $\vec\theta$.
This like~\eqref{eq:def RESS} tends towards $N$ for iid samples.

\paragraph{Efficiency}
Likewise, it is useful for practitioners to get a ball-park estimate of the \emph{efficiency} of a sampler:
\begin{align}
  \EFF := \frac{\RESS}{\nsamples} \in \R^+\,,
  \label{eq:def EFF}
\end{align}
If the number of samples per chain $\nsamples$ differs across chains, it is more appropriate to use the harmonic mean of $\nsamples$ than the mean; this ensures that $\EFF$ tends towards unity when samples are drawn iid from $\target$.
Although $\EFF$ is useful, $\RESS$ is more appropriate for comparisons between samplers.
Thinning can increase $\EFF$ without increasing estimation accuracy.

\paragraph{Normalization}
When looking at the distribution of sampler performance across examples it is more appropriate to look at normalized ESS ($\NESS$):
\begin{align}
  \NESS := \frac{\RESS}{\median_{\sampler \in \set{\sampler}}{\nsamples_\sampler}} \in \R^+\,,
\end{align}
where the median is taken across different samplers on the same example.
The $\RESS$, when evaluating with a fixed time limit, varies widely across examples as the computational cost of each sample varies greatly between benchmark examples.

\paragraph{ESS Deviation}
In order to evaluate the diagnostics in a meta-analysis we define the ESS deviation ($\ESSD$) metric which gives a sense on whether the ESS is biased or a generally poor predictor of estimation accuracy.
The $\ESSD$ is defined as:
\begin{align}
  \ESSD := \norminv\!\left(\chisq_\Nchains{\CDF}\left(\frac{\ESS}{\RESS} \Nchains\right)\right) \in \R\,,
  \label{eq:def ESSD}
\end{align}
where $\norminv(\cdot)$ is the inverse CDF of the standard normal.
ESSD has a standard normal distribution (under CLT assumptions) if the estimates are derived $\ESS$ iid samples; $\ESSD > 0$ indicates the estimation is higher error than expected from ESS.
More precisely, if $\hat{\theta}$ is derived from $m$ iid samples then,
\begin{align}
  \hat{\theta} &\approach \normal(\theta, \sqrt{R/m})
  \implies \sqrt{m/R} (\hat{\theta} - \theta) \sample \normal(0,1) \nonumber \\
  \implies &\sum_{k=1}^K \frac{m}{R} (\hat{\theta} - \theta)^2 = \frac{m}{\RESS}K \sample \chisq_K\,, \label{eq:deriv ESSD}
\end{align}
which implies that $\ESSD \sample \normal(0,1)$.
Note that~\eqref{eq:def ESSD} is merely a transformation to put the RESS-vs-ESS performance ratio on a standardized scale, which does not imply issues if the CLT assumption in~\eqref{eq:deriv ESSD} does not hold exactly.

\paragraph{Meta-analysis}
In our meta-analysis, we perform a Gaussian process regression to predict $\ESSD$ from $\ESS \in \R^+$, Gelman-Rubin $\GR \in [1,\infty)$, and Geweke $\G \in \R$.
We also include the dimension $\Dimx$ of the sample space $\vec x$.
Recall that if $\ESS$ is a perfect predictor of MCMC performance then ESSD will resemble white-noise (i.e., iid standard normal)\@.
Given that the scales of diagnostics vary widely, we use $\log\ESS$, $\log|\GR-1|$, and $\log|\G|$ to put them all on a sensible scale.

To assess the regression, we test on a held-out 20\% test set of unseen examples (i.e., we do random split on a per example basis) to see if we can predict the $\ESSD$ on new unseen benchmark examples from the MCMC diagnostics.
We compare performance of the regression to linear regression and an iid normal to see if the features provide any predictive gain.
Furthermore, we assess the predictive value of each feature by performing the regression after removing each feature and studying the performance delta.

\begin{table}
\centering
\caption{{\small Quantitative summary on sampler performance.
We show the NESS on various estimation tasks (e.g., $\mu$ vs $\sigma^2$) averaged over all examples on the left.
The right shows the probability of success, i.e., how often $\RESS \geq 12$.
The first three rows are different proposals for random walk Metropolis; and Mix is a compound proposal of NUTS and Gauss.
For both NESS and prob.~success, higher is better.}}
\label{tbl:overall perf}
{\footnotesize
\begin{tabular}{l|rrr|rrr}
\toprule
~ & \multicolumn{3}{c}{NESS} & \multicolumn{3}{c}{prob.~success} \\ \midrule
sampler               &  KS      &  $\mu$    &  $\sigma^2$ & KS & $\mu$ & $\sigma^2$ \\
\midrule
Cauchy  &          .004 &           .004 &          .003 &          .604 &          .582 &          .441 \\
Laplace &          .007 &           .004 &          .006 &          .566 &          .547 &          .439 \\
Gauss   &          .007 &           .005 &          .007 &          .585 &          .565 &          .436 \\ \midrule
HMC     &          .061 &           .151 &          .106 &          .580 &          .604 &          .531 \\
NUTS    & \textbf{.068} &  \textbf{.375} & \textbf{.115} &          .875 &          .783 &          .711 \\
emcee   &          .016 &           .038 &          .025 &          .389 &          .489 &          .379 \\
mix     &          .067 &           .164 &          .113 & \textbf{.911} & \textbf{.825} & \textbf{.715} \\
slice   &          .044 &           .078 &          .070 &          .745 &          .703 &          .643 \\
\bottomrule
\end{tabular}
}
\vspace{-4mm}
\end{table}

\begin{figure*}
    \centering
    \begin{subfigure}[b]{0.3\textwidth}
        \includegraphics[width=\textwidth]{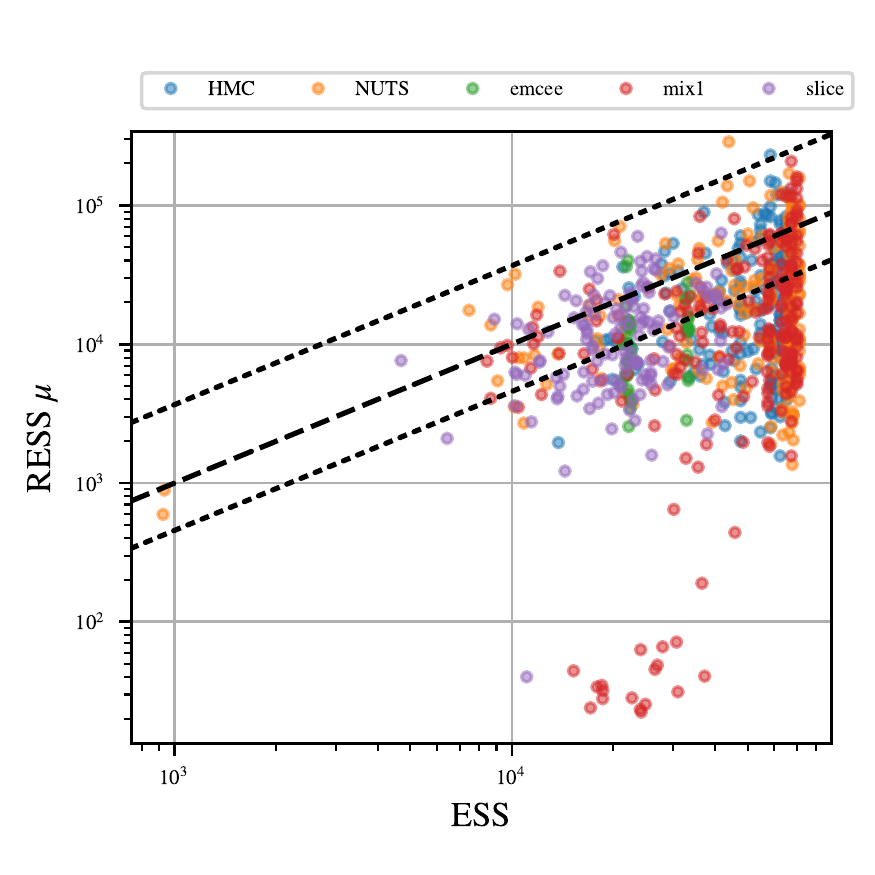}
    \end{subfigure}
    \begin{subfigure}[b]{0.3\textwidth}
        \includegraphics[width=\textwidth]{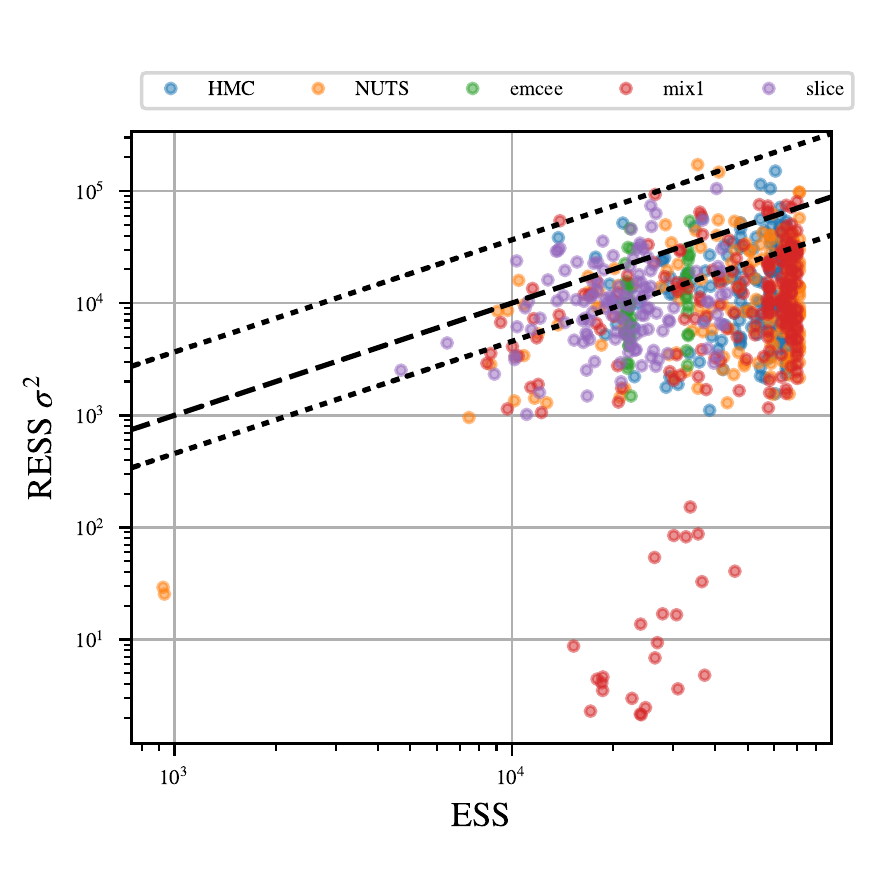}
    \end{subfigure}
    \begin{subfigure}[b]{0.3\textwidth}
        \includegraphics[width=\textwidth]{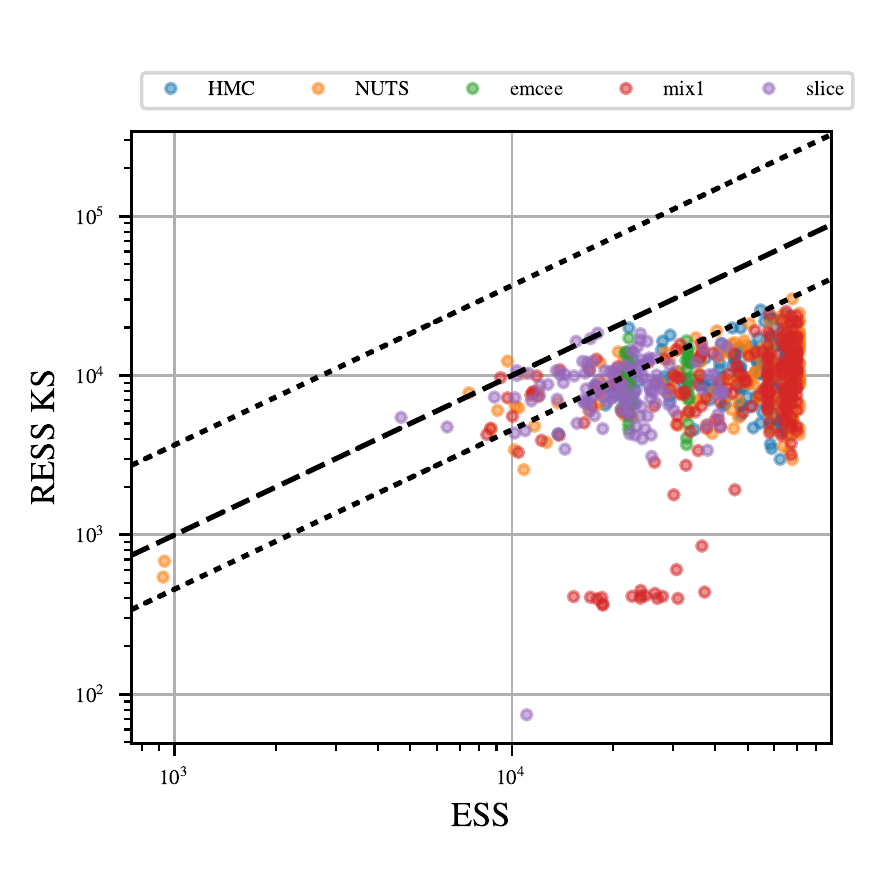}
    \end{subfigure}
    \caption{{\small Calibration plots of the ESS diagnostic against real ESS with $\hat{\theta}$ being the mean (left) or variance (center), and as well as the KS (right)\@.
    We show the diagonal for a perfect match in dashed black.
    In dotted black we show the 95\% region for what the observed real ESS would be if the estimates $\hat{\theta}$ were derived from ESS iid samples.
  The RESS is below the lower error bar 55\% of the time for for mean estimation, 68\% for variance, and 83\% for KS; these would be 2.5\% if a chain with $\ESS=m$ were functionally equivalent to $m$ iid samples.}}
  \label{fig:calibration}
\end{figure*}

\section{RESULTS}
\label{sec:results}

We first show an overall summary of final performance using NESS at the end of 15 minutes per chain, with $\Nchains=8$ chains in Table~\ref{tbl:overall perf}.
The box plots in Figure~\ref{fig:box} provide a sense of the variation.
We found the NESS of the samplers to generally be bimodal: either the samples achieve an efficiency above 1\% or they completely fail with an RESS $<$ 1.
Therefore, in Figure~\ref{fig:box} we show the box plots after excluding the complete failures.
Inspired by the rule of $N=12$ from~\citet{Mackay2003}, we use an RESS of 12 to threshold failure-vs-success.
Correspondingly, Table~\ref{tbl:overall perf} also provides an overall success probability for each method.
Emcee shows the most bimodal performance: while sometimes achieving a high NESS competitive with other advanced methods, it has the lowest success probability.
Emcee also has the lowest efficiency of any methods except random walk Metropolis, but makes up for its lack of efficiency with higher per sample speed.

Other results from Figure~\ref{fig:box} are unsurprising: NUTS and HMC are the highest performers, despite their higher per sample cost.
Slice sampling also makes a ``strong showing'' with its performance more competitive in the lower dimensional examples.
Random walk Metropolis methods generally have an efficiency in the 0.1\% to 1\% range while slice sampling and HMC based methods have efficiencies in the ball park of 2\% to 40\%, with NUTS showing the highest performance.
Emcee seems to vary widely.
Note that although the compound proposal (mix) does not substantially increase NESS (over NUTS), when the methods succeed Figure~\ref{fig:box}, mix increases the chance of success (Table~\ref{tbl:overall perf}).

\begin{table}
\centering
\caption{{\small Results of meta-analysis.
We show the MSE and log-loss of different models attempting to predict the ESSD for mean estimation on a held-out 20\% of unseen examples.
The log-loss has the advantage that it is parameterization invariant and provides the same results in $\ESSD$ or $\ESS$ space.
The GP- rows show the results of GP regression without the feature named.
GP shows the performance of the GP using all features.
We assess the statistical significance of the delta to GP using a pairwise t-test in p.}}
\label{tbl:meta results}
{\footnotesize
\begin{tabular}{|l|ll|ll|ll|}
\toprule
method &        {MSE} &     {p} &    {NLL (nats)} &     {p} \\
\midrule
GP     &   2.8588       &    {--} &   ~0             &    {--} \\
GP-D   &  \textbf{2.779(70)}    &  0.0252 &  \textbf{-0.0096(97)}    &  0.0504 \\
GP-ESS &  3.16(23)     &  0.0097 &   ~0.045(31)     &  0.0034 \\
GP-G  &  2.858(1)     &  0.0198 &  -0.0001(1)         &  0.0016 \\
GP-GR &   3.17(20)     &  0.0017 &   ~0.045(25)     &  0.0005 \\
iid     &   3.30(28)     &  0.0016 &   ~0.067(36)     &  0.0003 \\
linear  &  3.03(19)     &  0.0726 &   ~0.027(25)     &  0.0350 \\
\bottomrule
\end{tabular}
}
\vspace{-3mm}
\end{table}

We show calibration plots of ESS in Figure~\ref{fig:calibration} and efficiency in Figure~\ref{fig:box}.
The ESS diagnostic is clearly best calibrated for mean estimation, which is not surprising given it was derived for that purpose.
However, the ESS diagnostic clearly has an optimistic bias.
These results provide caution of ESS\@.

Finally, we present the results of the meta-analysis to predict ESS deviation.
We report the predictive value provided by various features in Table~\ref{tbl:meta results} by showing how much performance changes when they are removed.
ESS appears very predictive in Figure~\ref{fig:calibration}, but the relationship has already largely been accounted for with ESSD~\eqref{eq:def ESSD}.
In log-loss, the remaining predictive utility of ESS equals that of Gelman-Rubin.
Geweke and the dimension $D$ show no predictive utility.
Predictive performance of ESSD goes up when they are removed.

\vspace{-3mm}
\section{CONCLUSIONS}
\vspace{-3mm}

We have presented a general system to benchmark the real performance of MCMC samplers on realistic problems.
The data-driven nature of the benchmark makes it a highly novel development.
This benchmark is intended to become a general service that will become as wide spread as COCO or MLcomp.
Careful attention has been paid to fairly and sensibly derive metrics that compare samplers.
This benchmark will evolve with time by including ever more models in phase~1 and more advanced example densities in phase~2.

\bibliographystyle{abbrvnat}
\bibliography{benchmark_refs} 

\end{document}